\documentclass[runningheads]{llncs}
\usepackage[T1]{fontenc}
\usepackage[misc]{ifsym}
\usepackage{graphicx}
\usepackage{subfig}
\usepackage{amsmath}
\usepackage{amssymb}
\usepackage{amsfonts}
\usepackage{makecell}
\usepackage{multirow}
\usepackage{comment}
\usepackage{etoolbox}
\makeatletter
\patchcmd{\ps@headings}{\rlap{\thepage}}{}{}{}
\patchcmd{\ps@headings}{\llap{\thepage}}{}{}{}
\makeatother
\begin{document}

\title{
Bridging the Gap between Sparse Matrix Reordering and Factorization: A Deep Learning Framework for Fill-in Reduction
}
\titlerunning{A Deep Learning Framework for Fill-in Reduction}
%
\author{Ziwei Li\inst{1,2} \and
Tao Yuan\inst{1} \and
Shuzi Niu\inst{1}\textsuperscript{(\Letter)} \and
Huiyuan Li\inst{1}}

\institute{
Institute of Software, Chinese Academy of Sciences, Beijing, China\\
\email{\{liziwei2021,yuantao,shuzi,huiyuan\}@iscas.ac.cn}
\and
University of Chinese Academy of Sciences, Beijing, China}

\maketitle              
\begin{abstract}

Sparse matrix reordering can significantly reduce the fill-in during matrix factorization, thereby decreasing the computational and storage requirements in sparse matrix computations. Finding a minimal fill-in ordering is known to be an NP-hard problem. Moreover, there is a paradox: matrix reordering is applied before matrix factorization, but fill-ins that matrix reordering methods aim at are generated from matrix factorization. To bridge the gap between reordering and factorization, we propose a deep learning framework to minimize a fill-in surrogate function based on spectral embedding. 
First, we employ a multi-grid-like GNN architecture to learn to approximate the smallest eigenvectors of its graph Laplacian matrix, i.e. spectral embedding, and capture the global structural information of the matrix. Then, another multi-grid-like GNN architecture is used to minimize the potential space where fill-in can occur based on the rank distribution. Experimental results indicate that our approach achieves competitive performance compared with traditional graph-theoretic algorithms and deep learning methods.

\end{abstract}
\section{Introduction}
\begin{figure}
     \centering
    \captionsetup[subfloat]{labelsep=none,format=plain,labelformat=empty}
     \subfloat[$A$]{\includegraphics[width=0.125\linewidth]{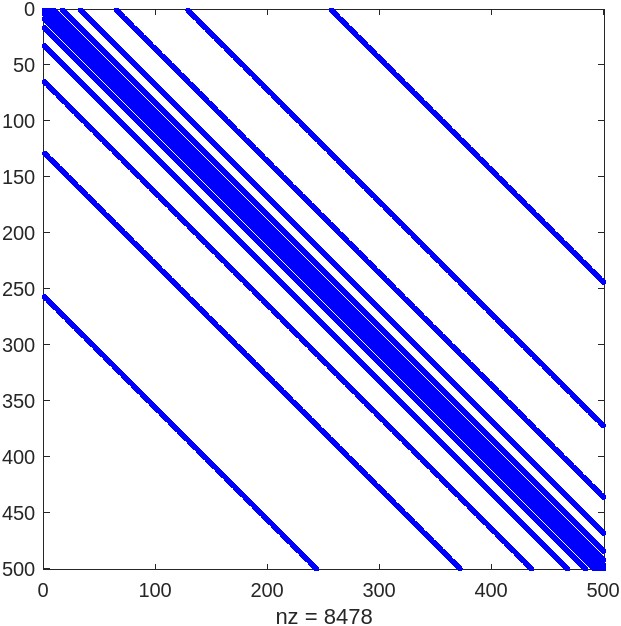}}
     \subfloat[$A_a$]{\includegraphics[width=0.125\linewidth]{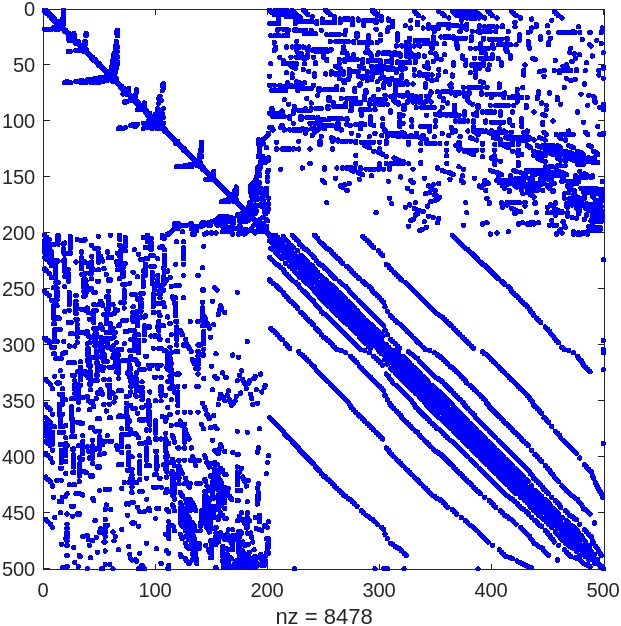}}
     \subfloat[$A_d$]{\includegraphics[width=0.125\linewidth]{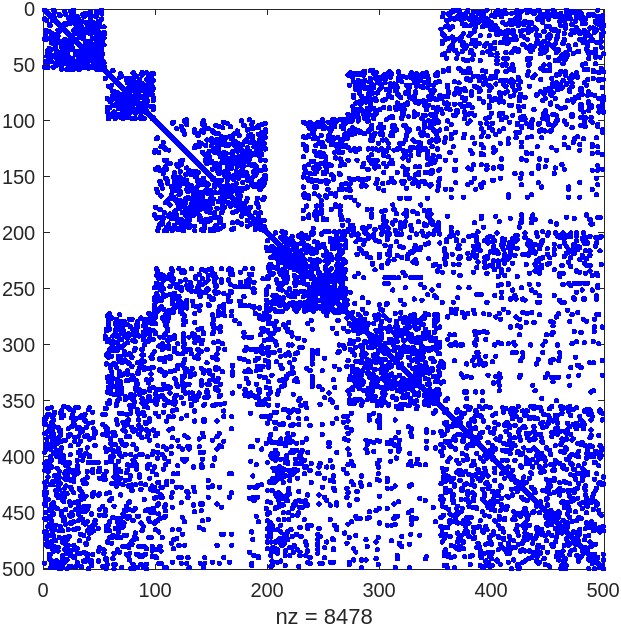}}
     \subfloat[$L+U$]{\includegraphics[width=0.125\linewidth]{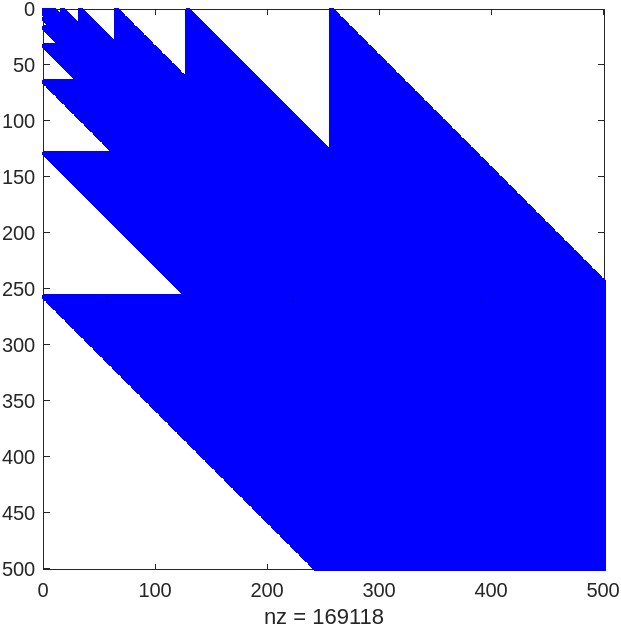}}
     \subfloat[$L_a+U_a$]{\includegraphics[width=0.125\linewidth]{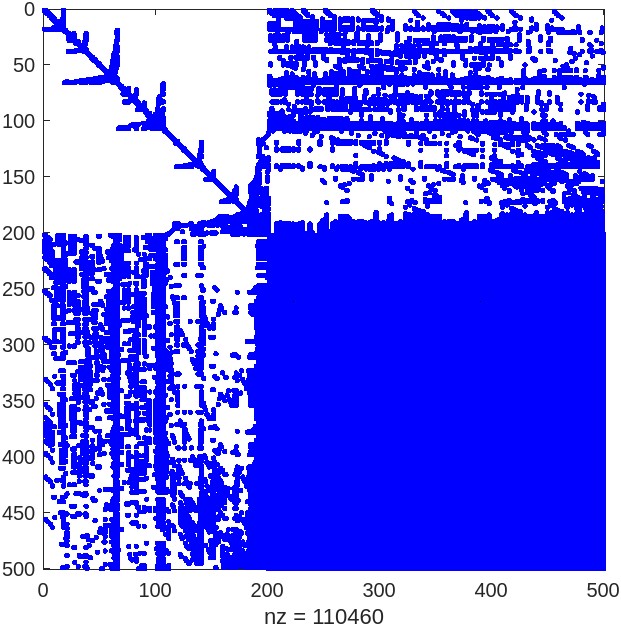}}
     \subfloat[$L_d+U_d$]{\includegraphics[width=0.125\linewidth]{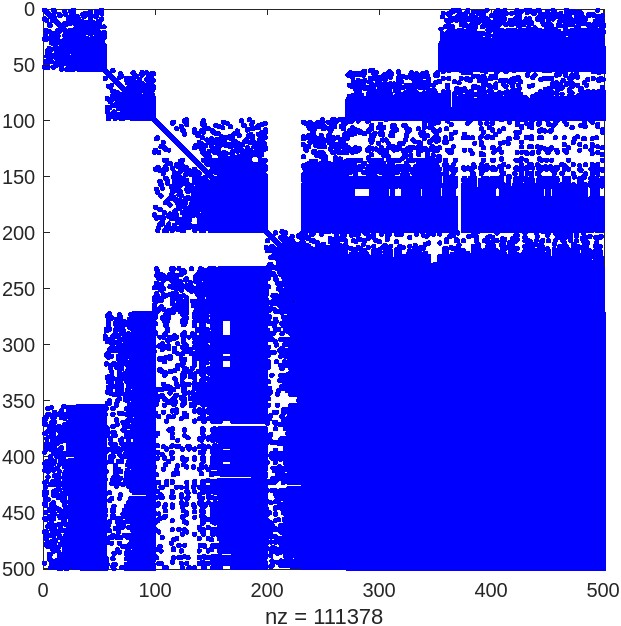}}
    \subfloat[Sparsity]{\includegraphics[width=0.125\linewidth]{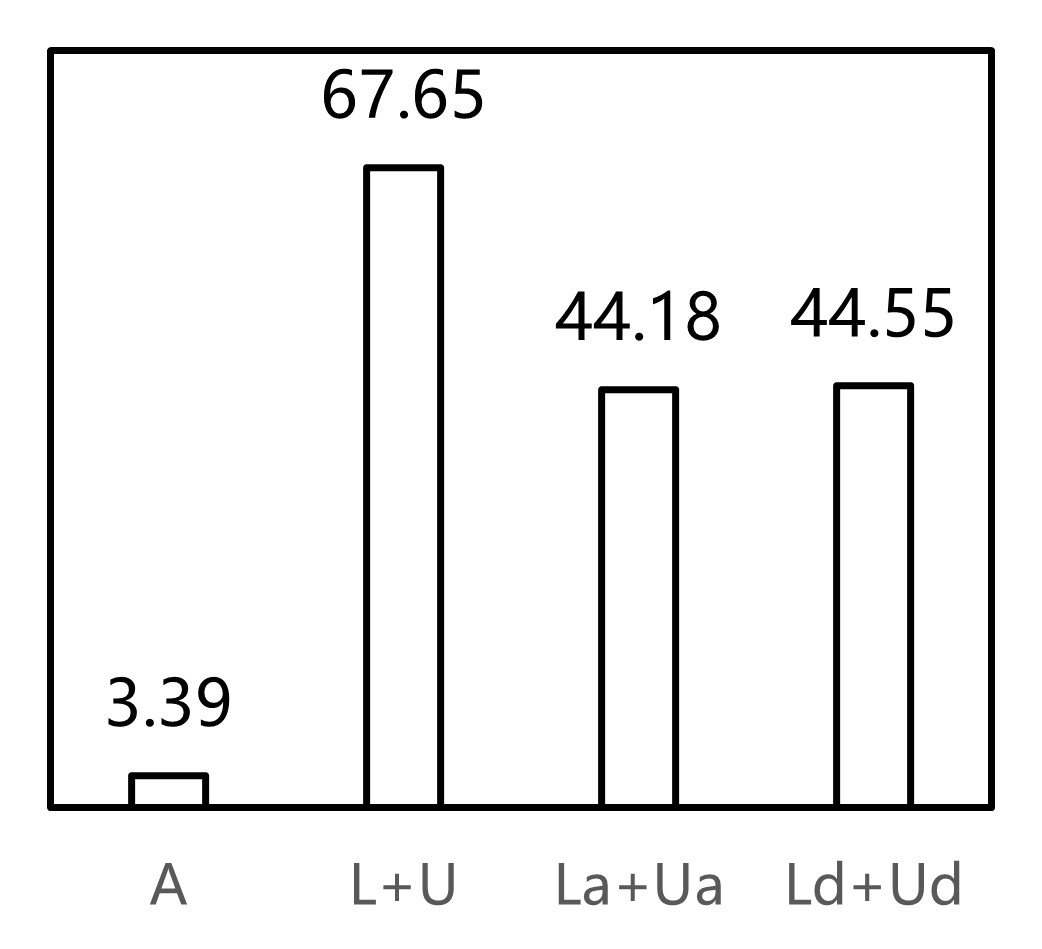}}
     \subfloat[Fill-ins]{\includegraphics[width=0.125\linewidth]{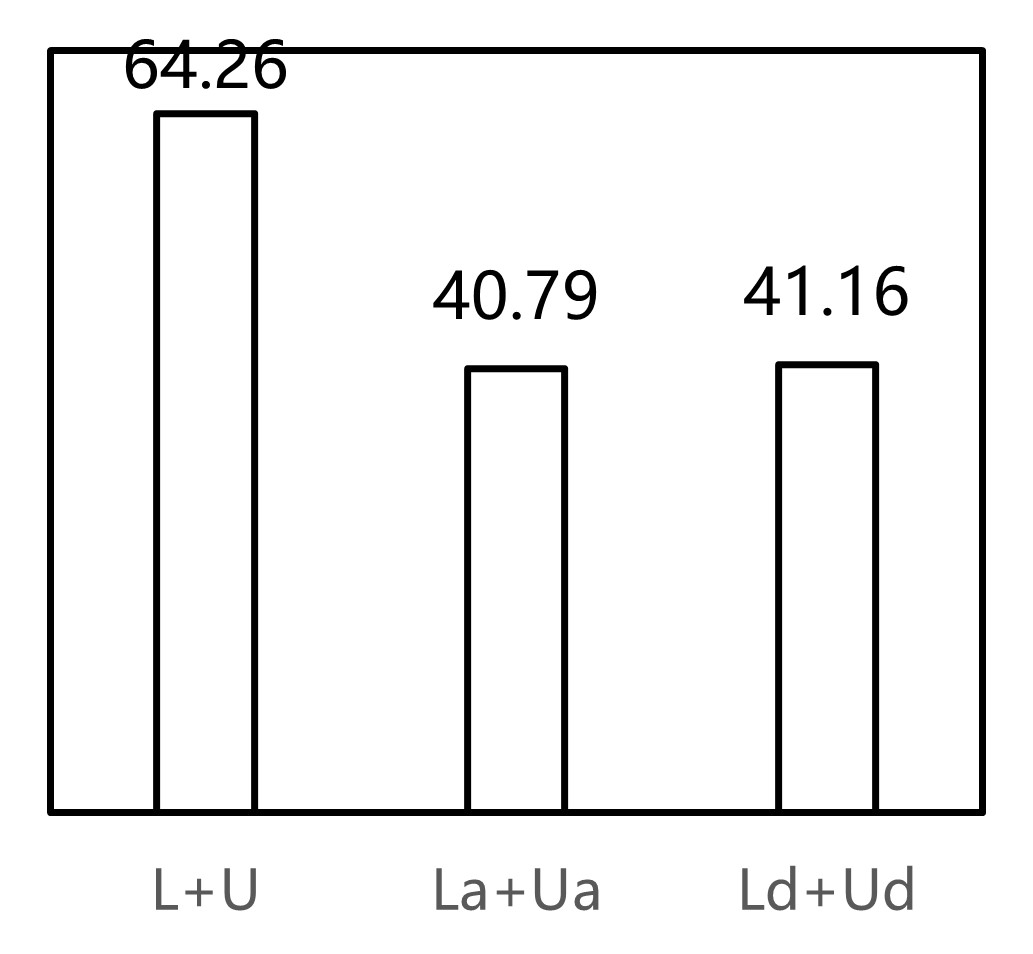}}
     \caption{LU factorization of the matrix Trefethen\_500 with different reordering methods generates different fill-ins. Trefethen\_500 without reordering and with AMD and Nested Dissection ordering are denoted as $A$, $A_a$ and $A_d$ separately. Each ordering is derived with MATLAB function as $r_a=\text{amd}(A)$,  $r_d=\text{dissect}(A)$. $A_*=A(r_*,r_*)$ and LU factorization is performed by $A_*=L_*U_*$.}
     \label{fig:fill-in}
 \end{figure}

In High Performance Computing (HPC), efficiently solving sparse linear systems is crucial for tackling complex real-world problems. Sparse matrices, common in large-scale applications, are characterized by a overwhelming majority of zero elements. Since only non-zero elements need to be stored and processed, exploiting this sparsity can lead to significant reductions in both storage and computational costs compared to dense matrices. Direct solution methods for sparse linear systems \cite{Duff1986Direct} typically involve four main steps: matrix reordering, symbolic factorization, numerical factorization, and back substitution. As the initial step, matrix reordering directly influences the efficiency of the entire solution process. By rearranging rows and columns to minimize fill-in, optimal ordering can substantially reduce resource usage in subsequent steps. Here fill-in means the number of non-zero entries that appear in positions initially occupied by zeros during factorization as show in Fig.~\ref{fig:fill-in}. Therefore, matrix reordering is essential for efficiently solving sparse linear systems in HPC.

A sparse matrix can be directly represented as an undirected or directed graph, where rows and columns represent nodes, and non-zero elements indicate edges. This graph-based representation effectively highlights the sparsity pattern by focusing solely on existing connections, facilitating the identification of structural patterns and enabling optimized computational strategies.
Sparse matrix reordering methods often convert the matrix into its corresponding adjacency graph and apply graph theoretical techniques to optimize node ordering. Specifically, the optimal matrix ordering problem can be formulated as finding a node ordering that minimizes additional edges generated during the graph elimination process. However, this problem is known to be NP-hard~\cite{Yannakakis1981}.

Most existing algorithms for sparse matrix reordering are heuristic, employing graph-theoretical approaches to find approximately optimal solutions. Representative methods include Reverse Cuthill-McKee (RCM)~\cite{RCM}, Nested Dissection (ND)~\cite{ND}, Approximate Minimum Degree (AMD)~\cite{AMD}, and Fiedler Ordering~\cite{spectral}. However, these methods cannot guarantee optimal solutions due to the NP-hardness of the fill-in minimization problem. Furthermore, computing the eigenvector required for Fiedler Ordering is computationally intensive.
Although some studies have attempted to use machine learning techniques to solve NP-complete problems such as the Travelling Salesman Problem (TSP)~\cite{TSP} and Vertex Cover~\cite{VertexCover}, few~\cite{AMG,GP,Acceleration,alphaelim} have focused on applying deep learning methods to systematically solve these graph-based problems common in HPC, aiming to accelerate sparse matrix computations.

Leveraging the advantages of deep learning methods, it is promising to cast the matrix reordering problem with deep graph neural networks. But it still faces the following two challenges. One is that the original problem provides nothing but graph structures. Even explicit node features, common in traditional AI applications, are absent. This challenge makes it more difficult to capture structural information. The other lies in the minimization of fill-in, that appears in the factorization step serves as the goal of matrix reordering, applied before factorization. This gap makes it hard to define a suitable loss function for training. Additionally, the number of non-zero elements introduced during LU decomposition is a discrete variable, which cannot be directly optimized through gradient-based methods.

To address these challenges, we propose an Unsupervised Direct Node Ordering (UDNO) method that directly predicts the node ordering of large graphs through a two-stage network to minimize fill-in. Our approach consists of two key steps. In the first stage, we employ a multi-level graph neural network architecture that leverages coarsening and interpolation techniques to approximate the eigenvectors corresponding to the first \(d-1\) smallest non-zero eigenvalues of the Laplacian matrix. These resulting eigenvectors are then employed as node features in the second stage. This graph representation learning approach mitigates challenges associated with missing node features in the adjacency graph, while avoiding the computational expense of exact spectral factorization. 

In the second stage, we introduce a novel loss function based on the 1-sum metric~\cite{spectral}, which quantifies the sum of position differences in the ordering between connected nodes. Inspired by the rank distribution~\cite{SoftRank}, we treat each node position as a random variable to adjust the inherently discrete and non-differentiable nature, and derive the expectation of 1-sum metric as the objective. Thus gradient based optimization methods like Adam can be used for training. This approach effectively circumvents the limitation of simply casting matrix reordering problem in supervised learning.
Experimental results show that our proposed method outperforms traditional graph-theoretic methods and existing deep learning techniques.

The major contribution of our proposed method lies in the following four aspects: (1) The introduction of the spectral embedding model, which provides a good initial embedding to avoid heuristic feature design; (2) the proposal of expected 1-sum metric as optimization objective to minimize the potential space where fill-in may occur; (3) Theoretical analysis of relationship among spectral embedding, fill-in and 1-sum metric; (4) Superior performance on benchmark sparse matrix collections.

\section{Problem Definition and Related Work}

This section begins by formally introducing sparse matrix reordering and its correspondence with graph representations. We then review the commonly used graph-theoretic methods for matrix reordering, as well as AI-based approaches.
\subsection{Formalization of Sparse Matrix Reordering}
To solve a sparse linear system \( A x = b \), where \( A \in \mathbb{R}^{n \times n} \) is a sparse matrix, we typically decompose \( A \) into the product of a lower triangular matrix \( L \) and an upper triangular matrix \( U \) (i.e., perform LU factorization) before solving. Matrix reordering seeks a permutation matrix \( P \in \mathbb{R}^{n \times n} \) such that the reordered matrix \( A' = P A P^T \) yields \( L \) and \( U \) that remain as sparse as possible after decomposition, thereby minimizing fill-in. The fill-in is quantified as \( L.nnz + U.nnz - A.nnz \), where \( L.nnz \), \( U.nnz \), and \( A.nnz \) represent the number of nonzero elements in \( L \), \( U \), and the original matrix \( A \), respectively. In the case of symmetric matrices, \( U = L^T \). Here, the LU decomposition is referred to as the Cholesky decomposition.

The symmetric matrix \( A \) can be represented by an undirected graph \( G = (V, E) \), where each node in \( V \) corresponds to a row or column of \( A \), and each edge in \( E \) represents a nonzero off-diagonal entry in \( A \). The Cholesky decomposition of the reordered matrix, which is equivalent to Gaussian elimination, corresponds to the node elimination process in the graph \( G \). At each node elimination step, new edges are created between all previously unconnected neighbors, corresponding to new nonzero elements generated during Cholesky decomposition.

Finding the permutation matrix \( P \) is equivalent to determining a permutation \( \pi : V \to \{1, 2, \dots, n\} \), where \( \pi(v) = j \) indicates that node \( v \) is eliminated at position \( j \). 
Thus, the row or column corresponding to node \( v \) in the original matrix becomes the \( j \)-th row or column in the reordered matrix. To construct the permutation matrix \( P \), we rearrange the rows of the identity matrix according to the permutation \( \pi \). Assume that each node \( v \in V \) is assigned an original elimination order index \( i \in \{1, 2, \dots, |V|\} \). The permutation \( \pi \) maps node \( v \) to a new elimination position \( \pi(v) = j \). Specifically, \( P \) is defined such that
\(
P[i, \pi(v)] = 1,
\)
with all other entries in each row being zero. An optimal node ordering significantly reduces the fill-in, thereby improving the efficiency of the entire solution process.
\subsection{Graph Theoretical Algorithms}

In the task of matrix reordering for reducing fill-in, various graph-theoretic approaches have been developed. 
Among these, the Cuthill-McKee (CM)~\cite{RCM} algorithm and its reverse variant, Reverse Cuthill-McKee (RCM)~\cite{RCM}, are well-known for reducing matrix bandwidth. The CM algorithm begins by selecting a pseudo-peripheral node and performing a breadth-first search on the adjacency graph, followed by sorting nodes in ascending degree order. The RCM algorithm further enhances this by reversing the order generated by CM, achieving a greater reduction in bandwidth.

Another class of matrix reordering algorithms is based on the degree of nodes in the graph representation.
The Minimum Degree (MD)~\cite{MD} algorithm iteratively selects the node with the lowest degree during elimination. 
Variants of this method include Multiple Minimum Degree (MMD)~\cite{MMD} and Approximate Minimum Degree (AMD)~\cite{AMD} among others. MMD extends the MD approach by selecting multiple nodes with the minimum degree simultaneously. AMD uses approximation techniques to simplify degree calculations, demonstrating strong performance in practical applications.

Additionally, graph partitioning methods, such as Nested Dissection (ND)~\cite{ND} and multilevel partitioning techniques~\cite{MGP}, are widely employed to reduce fill-in during matrix factorization. ND works by recursively partitioning the graph into smaller subgraphs through the removal of small sets of separator nodes. Multilevel partitioning methods, including multilevel recursive bisection~\cite{MRB} and multilevel \(k\)-way partitioning~\cite{kway}, enhance this approach by incorporating graph coarsening and refinement stages. 

Tools like METIS~\cite{metis} and SCOTCH~\cite{scotch} effectively implement both ND and multilevel partitioning strategies to generate efficient elimination orderings for sparse matrices. These tools reduce fill-in between subgraphs by minimizing interconnections through graph partitioning. Furthermore, they integrate reordering algorithms such as AMD to reorder nodes within each subgraph, thereby minimizing fill-in within the subgraphs themselves. 

Spectral ordering methods are also effective; they utilize the spectral properties of the graph Laplacian to reduce fill-in during matrix factorization. Specifically, Fiedler Ordering~\cite{spectral} reorders the matrix directly based on the values of the Fiedler vector's elements. By emphasizing the structural information captured by the Fiedler vector, this arrangement reduces the matrix bandwidth and envelope size, thereby decreasing the likelihood of fill-in occurring during the elimination process. However, computing the Fiedler vector is computationally expensive, particularly for large-scale graphs, which can limit the practical applicability of this method.

All the aforementioned methods are graph-theoretical heuristics that generally do not yield optimal solutions. While algorithms like AMD and RCM perform well on medium-sized matrices, their effectiveness declines as the matrix dimensions increase. Moreover, the efficiency of these heuristic algorithms heavily depends on the characteristics of the underlying graph, such as sparsity patterns and connectivity. Consequently, in cases where the graph structure is irregular or complex, these methods may fail to significantly reduce fill-in.

\subsection{AI-Based Methods for Sparse Matrix Reordering}
In recent years, there has been a growing interest in applying artificial intelligence (AI) techniques to optimize sparse matrix reordering, aiming to reduce fill-in during matrix factorization.

Gatti et al.~\cite{GP} proposed a method for graph partitioning and sparse matrix ordering utilizing reinforcement learning and graph neural networks. They employ an Advantage Actor-Critic (A2C) framework that iteratively partitions the graph by selecting small separators, which contributes to reducing the fill-in during matrix factorization. 
Booth et al.~\cite{Acceleration} introduced a technique to accelerate graph-based utility functions for sparse matrices using neural networks. Their approach involves training individual neural networks to predict the amount of fill-in that would result from applying various traditional ordering methods to a given sparse matrix. By predicting the fill-in for each ordering method—such as AMD or ND—the system can select the ordering algorithm that is expected to produce the least fill-in without having to perform a full factorization. 
Dasgupta et al.~\cite{alphaelim} introduced a deep reinforcement learning framework named Alpha Elimination to reduce fill-in during sparse matrix factorization. Their approach leverages Monte Carlo Tree Search to explore possible reordering sequences and uses a Convolutional Neural Network to approximate Q-values, representing the expected reward for each state-action pair. 

Although current AI methods have made significant progress in this field, several challenges remain. Gatti et al.'s work primarily focuses on optimizing graph partitioning rather than directly minimizing fill-in. Booth et al.'s research selects from existing methods but struggles to produce new orderings. Dasgupta et al.'s work, which applies dense convolutional networks, fails to fully leverage the sparse characteristics of the matrices.

\section{Method}

\begin{figure}[ht]
  \centering
  \includegraphics[width=0.95\linewidth]{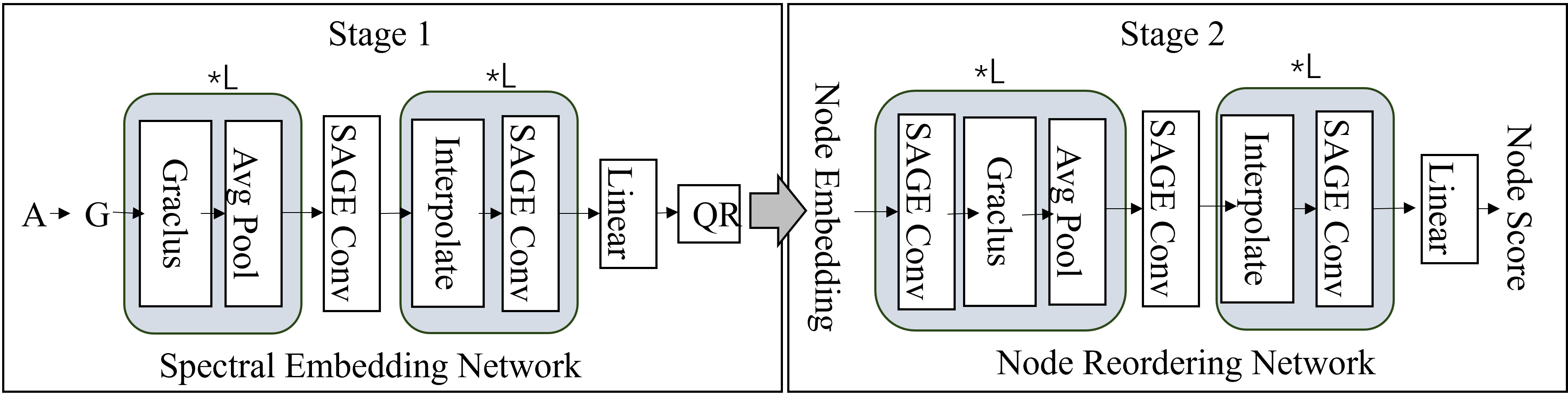}
  \caption{Architecture of UDNO.}
  \label{udno_architecture}
\end{figure}

We propose an Unsupervised Direct Node Ordering (UDNO) method based on a multi-grid-like GNN architecture, which predicts the ordering of all nodes in a graph simultaneously. UDNO operates in two stages: a spectral embedding stage and a node ordering stage.
In the spectral embedding stage~\cite{Spectral Embedding}, we transform the matrix into a graph and approximate the eigenvectors corresponding to the \( d-1 \) smallest nonzero eigenvalues of the graph Laplacian matrix to generate node embeddings, which are used as input for the second stage.
In the node ordering stage, we introduce a novel unsupervised envelope-based loss function to ensure that nodes connected or closely related are ordered as closely as possible in the final node ordering. Inspired by Softrank\cite{SoftRank}, we employ random variables to transform the originally non-differentiable objective into a continuous one.
The entire architecture is depicted in Fig.~\ref{udno_architecture}. A detailed explanation of the two stages is provided in the following sections. More details about code implementation are available at \url{https://github.com/plumvvvv/UDNO}.

\subsection{Spectral Embedding}
 
The first stage of our method adopts the setup of the initial phase of the graph partitioning work by Alice Gatti et al.~\cite{Spectral Embedding}.
This approach utilizes a multi-grid-like GNN architecture to approximate the \(d-1\) eigenvectors associated with the smallest non-zero eigenvalues, thereby serving as node-level feature representations of the graph. 

The process begins by repeatedly applying the Graclus clustering algorithm to the original graph \( G \), generating a sequence of coarsened graphs \( G^l \) at each level \( l = 0, 1, \dots, L_c \), until only two nodes remain in the final coarsest graph \( G^{L_c} \). Each clustering step produces an interpolation matrix \( I^l \) used in later stages. In the coarsest graph \( G^{L_c} \), the feature matrix \( F^{L_c} \) is defined, with each column vector representing the feature of one of the two nodes: \( [1, 0]^T \) and \( [0, 1]^T \), providing initial conditions for a natural separation in subsequent stages.

After coarsening, we alternate between SAGEConv layers (Sample and Aggregate convolution layers) and interpolation steps to reconstruct the original graph structure. The SAGEConv function updates node features through information exchange among neighboring nodes, following the update rule:
\begin{equation}
F^{l}_i = F^{l+1}_i W_1 + \frac{1}{|N(i)|} \sum_{j \in N(i)} F^{l+1}_j W_2,
\end{equation}
where \( F^{l}_i \) is the feature of node \( i \) at layer \( l \), \( N(i) \) represents its set of neighboring nodes, and \( W_1 \) and \( W_2 \) are learnable weight matrices shared across layers.
At each interpolation step, we apply convolutional layers followed by nonlinear activation functions, specifically the hyperbolic tangent (Tanh). 

Once the graph has been fully interpolated back to its original form, we apply additional linear layers to the feature matrix, updating it as \( F = F^{0} W + b \), where \( W \) and \( b \) are the learnable weight matrix and bias vector, respectively.
Finally, we perform QR factorization on the feature matrix to ensure orthogonal feature vectors with unit norms. 

The loss function is defined as follows:
\begin{equation}
Loss_1 = \left\| \tilde{L} \tilde{F} - \Lambda \tilde{F}  \right\|_2^2 + \sum_{i=1}^{d} \lambda_i,
\end{equation}
where \(\tilde{L}\) represents the normalized graph Laplacian, \(\tilde{F} = [f_1, f_2, \dots, f_d]\) represents the normalized eigenvectors predicted by the model, \(\Lambda = \text{diag}(\lambda_1, \lambda_2, \dots, \lambda_d)\) is a diagonal matrix containing the eigenvalues, and \(\lambda_i = f_i^T \tilde{L} f_i\) represents the approximate eigenvalue.

Here, we typically set \( d = 2 \), using only the Fiedler vector as node features for downstream task. For a sparse symmetric matrix \( A \in \mathbb{R}^{n \times n} \), the envelope is defined as
\(
\text{Env}(A) = \{ (i, j) \mid f_i(A) \leq j < i, \; i = 1, \dotsc, n \},
\)
where \( f_i(A) = \min\{ j \mid a_{ij} \neq 0, \; j \leq i \} \).
The two related envelope quantities, the 1-sum and 2-sum, are defined as:
\begin{equation}
\sigma_1(A) = \sum_{i=1}^n \sum_{j \in \text{row}(i)} (i - j), \quad \sigma_2^2(A) = \sum_{i=1}^n \sum_{j \in \text{row}(i)} (i - j)^2,
\end{equation}
where \( \text{row}(i) = \{ j \mid a_{ij} \neq 0, \; 1 \leq j \leq i \} \).

The values of \( \sigma_1(A) \) and \( \sigma_2^2(A) \) are closely related to the ordering of matrix \( A \). Minimizing these quantities restricts the positions where new non-zero entries can appear during matrix factorization, thereby reducing the potential range of fill-in and improving computational efficiency.
The second eigenvector of the Laplacian matrix of a connected graph \( G \), known as the Fiedler vector, is the optimal solution to the relaxed problem of minimizing \( \sigma_2^2(A) \).

The Fiedler vector is also widely utilized in spectral methods for nested dissection~\cite{spectral nested dissection} and in the partitioning of finite element meshes~\cite{Pothen1990,Simon1991}. It effectively serves as a compressed representation of matrix information by projecting the high-dimensional structure onto a one-dimensional space, providing valuable insights for matrix reordering.

\subsection{Node Reordering}

In the node reordering stage, the network architecture largely mirrors the first phase but incorporates an additional \texttt{SAGEConv} layer before each graph coarsening step. Unlike the first phase, this stage omits QR factorization and instead outputs a score vector, assigning each node \( u \) a score \( f(u) \) that reflects its position in the reordered graph.

We introduce a novel loss function aimed at optimizing the envelope-related quantity \( \sigma_{1,\min}(A) \). This problem can be re-expressed in terms of the graph \( G(V, E) \) corresponding to a sparse symmetric matrix \( A \) as follows:
\begin{equation}
\min_{\pi} \sum_{(u,v) \in E} |\pi(u) - \pi(v)|,
\end{equation}
where \( \pi(u) \) denotes the position of node \( u \) in the elimination ordering \( \pi \).

When the graph is augmented with the fill-in edges produced during its elimination process, the resulting graph becomes chordal. In Yannakakis's work~\cite{Yannakakis1981}, it is shown that \( \sigma_{1,\min}(A) \) is closely related to the number of edges (fill-in) that need to be added to \( G \) to obtain a chordal graph derived from its bipartite representation. Moreover, minimizing \( \sigma_{1,\min}(A) \) effectively reduces the potential space where fill-in can occur during the matrix factorization process.

To facilitate optimization, we approximate this discrete function with a continuous one using random variables, which serves as our loss function.
Assuming that node rankings start from zero, the ranking of node \( u \) is based on the number of other nodes with higher scores. Using an indicator function, it is expressed as:
\begin{equation}
\pi(u) = \sum_{u' \in V - \{u\}} \mathcal{I}(f(u) < f(u')).
\end{equation}
However, when node scores are deterministic values, the loss function becomes non-differentiable. Inspired by the work on SoftRank \cite{SoftRank}, instead of treating each node's score as a fixed value, we model it as a random variable \(S_u\) following a normal distribution with mean \( f(u) \) and variance \( \sigma^2 \). At this point, the rank of node \( u \) also becomes a random variable, denoted as
\begin{equation}
R_u = \sum_{u' \in V - \{u\}} \mathcal{I}(S_u < S_{u'}).
\end{equation}
Therefore, our loss function can be expressed as:  
\begin{equation}
\text{loss}_2 = E\left(\sum_{(u,v) \in E} |R_u - R_v|\right).
\end{equation}

Here, we can consider \( R_u \) as a Binomial-like random variable. Each comparison between node \( u \) and another node \( u' \) can be viewed as an individual trial, and there are \( N - 1 \) such trials in total. 
We then define \( p_{uu'} \) as the probability that node \( u \) ranks after node \( u' \), which is given by
\begin{equation}
p_{uu'} = P(S_{u'} - S_u > 0) = \int_0^{\infty} \mathcal{N}(f(u') - f(u), 2\sigma^2) \, ds.
\end{equation}
However, unlike a traditional binomial distribution where the success probability is constant, the success probability \( p_{uu'} \) varies for each comparison. Consequently, the actual distribution of \( R_u \) is more complex and lacks an explicit expression.

Based on the Central Limit Theorem, when \( N \) is sufficiently large, a binomial distribution with success probability \( p \) can be approximated by a normal distribution with mean \( N \cdot p \) and variance \( N \cdot p \cdot (1 - p) \). Although \( R_u \) is not a true binomial distribution due to varying success probabilities for different events, simulation results suggest that \( R_u \) approximately follows a normal distribution with mean \( \mu_{ru}=\sum_{u' \in V - \{u\}} p_{uu'} \) and variance \( \sigma_{ru}^2=\sum_{u' \in V - \{u\}} p_{uu'} \cdot (1 - p_{uu'}) \).

Since \( R_u \sim \mathcal{N}(\mu_{ru}, \sigma^2_{ru}) \) and \( R_v \sim \mathcal{N}(\mu_{rv}, \sigma^2_{rv}) \) are approximately independent normal distributions, we can compute the expected absolute difference \( E(|R_u - R_v|) \). Let \( R = R_u - R_v \); then \( R \) follows a normal distribution with mean \( \mu_R = \mu_{ru} - \mu_{rv} \) and variance \( \sigma^2_R = \sigma^2_{ru} + \sigma^2_{rv} \), i.e.,
\begin{equation}
R \sim \mathcal{N}(\mu_{ru} - \mu_{rv}, \sigma^2_{ru} + \sigma^2_{rv}).
\end{equation}
The probability density function (PDF) of the folded normal distribution \( |R| \) is given by:
\begin{equation}
f_{|R|}(y) = \frac{1}{\sqrt{2 \pi \sigma^2_R}} \left( e^{-\frac{(y - \mu_R)^2}{2 \sigma^2_R}} + e^{-\frac{(y + \mu_R)^2}{2 \sigma^2_R}} \right), \quad y \geq 0.
\end{equation}
Using this PDF, the expectation \( E(|R_u - R_v|) \) can be derived as:
\begin{equation}
\sqrt{\sigma^2_{ru} + \sigma^2_{rv}} \cdot \sqrt{\frac{2}{\pi}} \exp\left(-\frac{(\mu_{ru} - \mu_{rv})^2}{2(\sigma^2_{ru} + \sigma^2_{rv})}\right) 
+ (\mu_{ru} - \mu_{rv}) \left(1 - 2 \Phi\left(-\frac{\mu_{ru} - \mu_{rv}}{\sqrt{\sigma^2_{ru} + \sigma^2_{rv}}}\right)\right),
\end{equation}
where \( \Phi \) denotes the cumulative distribution function of the standard normal distribution.
By transforming the original discrete loss function into a continuous one using random variables, we have derived the final expression. This allows us to apply continuous optimization techniques to efficiently compute the loss function.

\section{Experiments}
To verify the effectiveness and generalization of our proposed method, we conduct comprehensive experiments on well designed training set and evaluate our models on a large collection of real-world benchmark sparse matrices. 

\subsection{Setting}
The sparse matrix collection is often originated from three methods. One is the existing public benchmark set, i.e. SuiteSparse Matrix collection~\cite{suitesparse}. It includes about $2,700$ sparse matrices from scientific computing and engineer applications, such as Computational Fluid Dynamic and structural problem. The matrix size, the number of rows or columns of a matrix, varies from tens to more than billions. Matrices from the 2D and 3D discretized problem are used for training. The other two methods are to generate Delaunay and FEM triangular meshes within different geometries. These generators are often used in scientific computing. For example, random Delaunay mesh method produces triangular meshes on square regions within $[0,1]^2$ and rectangular regions within $[0,2]\times[0,1]$. FEM triangular meshes are constructed with different geometries, such as GradeL, Hole3 and Hole6.

We construct two training sets for two stages respectively by mixing matrices from SuiteSparse~\cite{suitesparse}, Delaunay and FEM triangular mesh methods according to construction methods~\cite{Spectral Embedding} to increase the diversity of matrix sparsity patterns. Generated training matrices for phase 1 include $5,000$ matrices, with matrix size ranging from $100$ to $5,000$. Training matrices for phase 2 are composed of $100$ matrices, with matrix size ranging from $100$ to $500$.
For test matrix collection, we randomly selected $150$ matrices with matrix size exceeding $10,000$ from the SuiteSparse Matrix Collection~\cite{suitesparse}, denoted as SuiteSparse150. There are $44$, $25$, $12$ selected matrices from Structural problem, Computational Fluid Dynamics, and 2D and 3D Discretized problem separately, denoted as M$_{\text{SP}}$, M$_{\text{CFD}}$ and M$_{\text{2D3D}}$. M$_{\text{other}}$ represents the remaining $69$ selected matrices.

Both the spectral embedding and node ordering network architecture follow a multi-grid-like GNN~\cite{Spectral Embedding}. For spectral embedding, only non-parameterized clustering pooling is used. The initial and output dimension of the SAGEConv layer over the coarsest graph are $2$ and $32$. Both the input and output dimension of SAGEConv layer for each uncoarsened graph are $32$. For node ordering, the initial embedding dimension is $1$. The input and output dimension of two SAGEConv layers for each coarsened or uncoarsened graph are $16$ and $16$ respectively. The spectral embedding network is trained first with Adam optimizer. The learning rate is chosen from $0.00001$ to $0.1$ with best choice as $0.0001$. Parameters of the node ordering network are optimized with our proposed loss by freezing spectral embedding network parameters. They are also learned with Adam with learning rate $0.0001$. Moreover, the noise tolerance $s$ in the proposed rank loss is chosen as $0.0001$ from $0.00001$ to $0.01$.

The evaluation pipeline is performed like Fig.~\ref{fig:fill-in}. Given a matrix $A$, an ordering or permutation $r$ is computed from each baseline or our proposed method. The reordered matrix $A_*$ is derived as $A(r_*,r_*)$ or $P_*AP^T_*$, where \( P_* \) is the permutation matrix constructed by permuting the rows of the identity matrix \( I \) according to the permutation vector \( r_* \). It is fed into python interface $splu$ of SuperLU to obtain LU factorization $A_*=L_*U_*$. The number of fill-in generated during this factorization is measured as $L_*.nnz+U_*.nnz-A.nnz$. Ignoring the matrix size influence, we normalize the fill-in number as nnz\_ratio as below:
\begin{equation}
\text{nnz}\_{\text{ratio}}=\frac{L_*.\text{nnz}+U_*.\text{nnz}-A.\text{nnz}}{A.\text{nnz}}.
\end{equation}
The start and end timestamp for $splu$ are denoted as $t_{begin}$ and $t_{end}$, and the execution time for LU factorization is equal to $t_{end}-t_{begin}$.

\begin{table}[!htbp]
\centering
\caption{Performance Comparison across various ordering methods on the test matrix collection}\label{tab:effect}
\subfloat[nnz ratio]{
\begin{tabular}{|c|p{0.95cm}p{0.95cm}|p{0.95cm}p{0.95cm}|p{0.95cm}p{0.95cm}|p{0.95cm}p{0.95cm}|p{0.95cm}p{0.95cm}|p{0.95cm}p{0.95cm}|}
\hline
  & \multicolumn{2}{c|}{M$_{\text{SP}}${\tiny$\sim 29\%$}}
  & \multicolumn{2}{c|}{M$_{\text{CFD}}${\tiny$\sim 17\%$} }
  & \multicolumn{2}{c|}{M$_{\text{2D3D}}${\tiny$\sim 8\%$}} &\multicolumn{2}{c|}{M$_{\text{Other}}${\tiny$\sim 46\%$}} 
  & \multicolumn{2}{c|}{overall}\\
  \cline{2-11}
  &mean&dev&mean&dev&mean&dev&mean&dev&mean&dev\\
 \hline
\textbf{Natrual} &185.60& 	502.91 	&361.23 	&175.32  &474.71 &834.19 &493.29 	&944.26 &379.54 &	774.95 \\
\hline
\textbf{AMD}&169.24 &497.53 &378.22 &193.12  &528.85 &693.29 &417.09 	&651.80 	&346.85 &	589.88 \\
\textbf{Metis}&53.39 	&94.05 	&73.49 	&34.75 	 &82.29 	&73.32 	&\textbf{139.13} 	&\textbf{196.13} 	&98.49 	&150.53 \\
\textbf{Fiedler}&48.39 	&76.80 	&\textbf{56.42} 	&50.68 	 &\textbf{78.23} 	&65.95 	&149.72 	&221.16 	&98.73& 	164.33 \\
\hline
\textbf{$\overline{\text{Fiedler}}$}&55.66 	&93.92 	&61.62 	&35.23 	 &79.95 	&63.16 	&143.20 	&217.16 	&98.86 	&162.90 \\
\textbf{GPCE}&49.85 	&86.40 	&61.61 	&37.10 	 	&84.31 	&\textbf{59.75} 	&146.02 	&201.38 &100.60 &	154.90  \\
\hline
\textbf{UDNO}&\textbf{46.67} 	&\textbf{68.42} 	&59.57 	&\textbf{27.68}  &79.75 	&64.74 	&139.77 &199.05 	&\textbf{94.29} &	\textbf{148.43} \\
\hline
\end{tabular}
}
\\
\subfloat[LU factorization time (second)]{
\begin{tabular}{|c|p{0.95cm}p{0.95cm}|p{0.95cm}p{0.95cm}|p{0.95cm}p{0.95cm}|p{0.95cm}p{0.95cm}|p{0.95cm}p{0.95cm}|p{0.95cm}p{0.95cm}|}
\hline
  & \multicolumn{2}{c|}{M$_{\text{SP}}${\tiny$\sim 29\%$}}
  & \multicolumn{2}{c|}{M$_{\text{CFD}}${\tiny$\sim 17\%$} }
  & \multicolumn{2}{c|}{M$_{\text{2D3D}}${\tiny$\sim 8\%$}} &\multicolumn{2}{c|}{M$_{\text{Other}}${\tiny$\sim 46\%$}} 
  & \multicolumn{2}{c|}{overall}\\
    \cline{2-11}
  &mean&dev&mean&dev&mean&dev&mean&dev&mean&dev\\
 \hline
\textbf{Natrual} & 914 &	3198 &	467 &	1354 &	735 &	1610 &	603 &	2070 &	682 &	2300  \\
\hline
\textbf{AMD}&389 &	1043 &	1093 	&735 &	2361 &	3744 &	1587 	&3860 &	1251 &	3139 \\
\textbf{Metis}&217 	&545 &	46 &	252 &	36 &	81 &	102 &	232 &	121 &	342   \\
\textbf{Fiedler}&75 &	280 &	20 	&707 &	19& 	47& 	102 &	248 &	74 &	229   \\
\hline
\textbf{$\overline{\text{Fiedler}}$}&81 	&185 &	17 &	268& 	14 &	26 	&114 &	261 &	80 &	207 \\
\textbf{GPCE}&56 &	179 &	21 &	371 &	13 &	22 &	98 &	207 &	67 	&174 \\
\hline
\textbf{UDNO}&\textbf{52} &	\textbf{114} &	\textbf{16} &	\textbf{229} & \textbf{12} &	\textbf{22} 	&\textbf{98} &	\textbf{224} &	\textbf{64} 	&\textbf{168}  \\
\hline
\end{tabular}
}
\end{table}

There are two kinds of baselines: (1) Graph Theoretical methods include AMD~\cite{AMD}, Metis~\cite{metis} and Fiedler Vector~\cite{spectral}. (2) Deep Learning methods contain the spectral embedding denoted as $\overline{\text{Fiedler}}$, which is an approximate of Fiedler vector, and Naive Graph Neural Networks trained with supervision information denoted as \textbf{GPCE}. The spectral embedding is fed into a network of two SAGEConv layers, whose parameters are trained to approximate the best ordering among AMD, Metis and Fiedler vector on the same dataset as ours. Each node pair order is defined as Bradley Terry model, and the approximation is defined as the cross entropy between the best and predicted ordering in terms of node pair classification, i.e. \textit{P}airwise \textit{C}ross \textit{E}ntropy loss.
GPCE is used as a deep learning baseline method due to the lack of open-source deep learning baselines directly optimizing fill-in~\cite{alphaelim}. Other deep learning approaches related to matrix reordering either cannot generate new orderings~\cite{Acceleration} or focus on graph partitioning without aiming to reduce fill-in as the primary objective~\cite{GP}. Natural method means no matrix reordering in the following performance tables.

\subsection{Effectiveness Study}
Graph theoretical methods, like AMD, Metis and Fiedler, compute the ordering directly for each matrix in the test collection SuiteSparse150. Deep Learning methods, like $\overline{\text{Fiedler}}$, GPCE and UDNO, are first trained on the mixed matrices, and then predict their orderings in SuiteSparse150 with the learned weight parameters. Each ordering is fed into the evaluation pipeline mentioned above and its corresponding fill-in is measured in terms of nnz\_ratio and LU factorization time with second as the unit. For each problem group with more than $10$ matrices, the performance is shown in Table~\ref{tab:effect}.

UDNO achieves overwhelming performance compared with graph theoretical methods, i.e. AMD, Metis and Fiedler, according to the overall performance in Table~\ref{tab:effect}(a). The nnz\_ratio improvement of UDNO is about $4$ compared with the best baseline. Though UDNO is the second best on some problem group, like M$_{2D3D}$ and M$_{Other}$, its difference is small from the best, such as $1$ on M$_{2D3D}$ and $0.5$ on M$_{Other}$. The small nnz\_ratio gap disappears in terms of LU factorization time correspondingly in Table~\ref{tab:effect}(b). As far as we know, it is a first attempt to directly optimize the fill-in with deep learning methods in the unsupervised learning setting. In terms of both metrics, UDNO shows the promising results of deep learning methods for fill-in reduction.

UDNO is consistently better than deep learning baselines, i.e. $\overline{\text{Fiedler}}$ and GPCE, among all the grouped matrices and overall matrices listed in Table~\ref{tab:effect}(a). With $\overline{\text{Fiedler}}$'s output as the initial embedding, the significant improvement of UDNO shows its necessity of node ordering model. With the same initial embedding, GPCE's nnz\_ratio improvement is not so obvious compared with $\overline{\text{Fiedler}}$. These comparison results suggest the well design of network architecture and loss function in node ordering model. The detailed analysis will be in the ablation study.

Both $\overline{\text{Fiedler}}$ and GPCE show modest performances due to their high dependency on a certain or more graph theoretical methods in Table~\ref{tab:effect}(a) and (b). $\overline{\text{Fiedler}}$ is an approximation of Fiedler vector. Supervision information for learning GPCE's parameters is chosen among these three methods. Generally, the graph theoretical algorithms obtain the approximately optimal fill-in ordering. Thus the ordering derived from these deep learning methods is usually bounded by the ordering quality of its related graph theoretical methods.

%
Mean results of both nnz\_ratio and LU factorization time show the superior effectiveness of the proposed method. 
Besides, standard deviation in the table are also consistently lower than baseline methods in most cases. Even when UDNO's mean nnz\_ratio for M$_{\text{CFD}}$ is slightly higher than the best nnz\_ratio, its standard deviation is still the lowest among baselines. The lower deviation of UDNO shows its robustness to different kinds of matrices. We will explore its scalability further to different matrices in the following subsection.

\begin{table}[!htbp]
\centering
\caption{Nnzratio and LU factorization time with different loss functions}\label{tab:ablationloss}
\begin{tabular}{|c|p{0.8cm}p{0.8cm}|p{0.8cm}p{0.8cm}|p{0.8cm}p{0.8cm}|p{0.8cm}p{0.8cm}|p{0.8cm}p{0.8cm}|p{0.8cm}p{0.8cm}|p{0.8cm}p{0.8cm}|}
\hline
& \multicolumn{6}{c|}{nnz\_ratio}
  & \multicolumn{6}{c|}{LU factorization time $t_{LU}$ (second) }\\
\cline{2-13}
  & \multicolumn{2}{c|}{M$_{\text{SP}}$}
  & \multicolumn{2}{c|}{M$_{\text{CFD}}$ }
  & \multicolumn{2}{c|}{M$_{\text{SP,CFD}}$}
  & \multicolumn{2}{c|}{M$_{\text{SP}}$}
  & \multicolumn{2}{c|}{M$_{\text{CFD}}$ }
  & \multicolumn{2}{c|}{M$_{\text{SP,CFD}}$}\\
   \cline{2-13}
  &mean&dev&mean&dev&mean&dev&mean&dev&mean&dev&mean&sdev\\
 \hline
\textbf{$\overline{\text{Fiedler}}$} &55.66& 	93.92 	&61.62 	&52.89& 	57.82& 	81.08& 	80.63& 	185.19 	&17.14 &	30.98 &	57.63 &	151.56\\
\textbf{UDNO-1}&123.1
&	96.3&-&-&-&-&219.0
&	318.9&-&-&-&-\\
\hline
\textbf{PCE}&51.48 	&78.15 	&59.80 &	49.97 &	54.46 &	69.08 &	64.61 &	146.74 	&16.25 &	29.92 &	47.29 &	120.67 \\
\hline
\textbf{UDNO}&46.67 &	68.42 &	59.57 	&52.34& 	51.35 &	62.98 &	51.92 &	113.97 &	16.36 &	30.62 	&39.04 	&94.02 \\
\hline
\end{tabular}
\end{table}

\subsection{Ablation Study}
Introduction of spectral embedding model to derive the initial embedding and fill-in surrogate objective design without supervision information are two major points here. To explore the necessity of the spectral embedding model, we use the random initialization instead of spectral embedding and only keep phase 2, denoted as UDNO-1 in Table~\ref{tab:ablationloss}. The approximate Fiedler vector learned from spectral embedding model in phase 1 is denoted as $\overline{\text{Fiedler}}$, meaning UDNO-2.

Without the spectral embedding model, UDNO-1 performs the worst in Table~\ref{tab:ablationloss}. LU factorization based on reordering from UDNO-1 takes twice as long as $\overline{\text{Fiedler}}$. So we only evaluate nnz\_ratio in M$_{\text{SP}}$. Without the node reordering module, $\overline{\text{Fiedler}}$ achieve a little worse performance than UDNO. Thus, spectral embedding model plays a critical role in UDNO.

Different from traditional ranking loss or classification loss, our designed loss function is better for fill-in reduction reordering. To verify its effectiveness, we compare different loss functions with the same backbone as UDNO in Table~\ref{tab:ablationloss}. UDNO without the node reordering module is $\overline{\text{Fiedler}}$. Replacing the UDNO's loss function with Pairwise Cross Entropy mentioned above is denoted as PCE.

Compared with the Fiedler score estimated from spectral embedding model, $\overline{\text{Fiedler}}$ , PCE performs better in terms of nnz\_ratio and LU factorization time on matrices from both structural and computational fluid dynamics problem in Table~\ref{tab:ablationloss}. The result indicates the positive role of ranking loss in solve the general learning to ranking problem. However, it is not the best choice yet for the specific fill-in reduction reordering problem. Compared with UDNO, the performance improvement of PCE is modest. Thus it is necessary to incorporate the current loss to achieve the performance gain for UDNO. Another advantage is that UDNO loss is computed without any supervision information compared with PCE.
\begin{figure}[!htbp]
    \centering
    \subfloat[nnz ratio]{
    \includegraphics[width=0.5\linewidth]{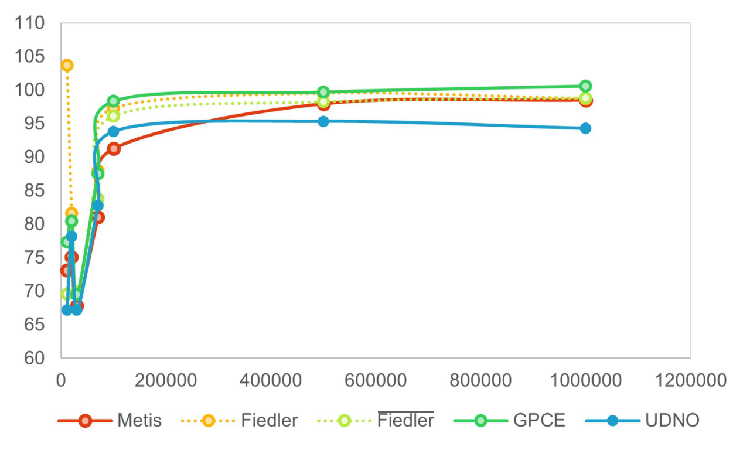}
    }
        \subfloat[ordering time (second)]{
    \includegraphics[width=0.5\linewidth]{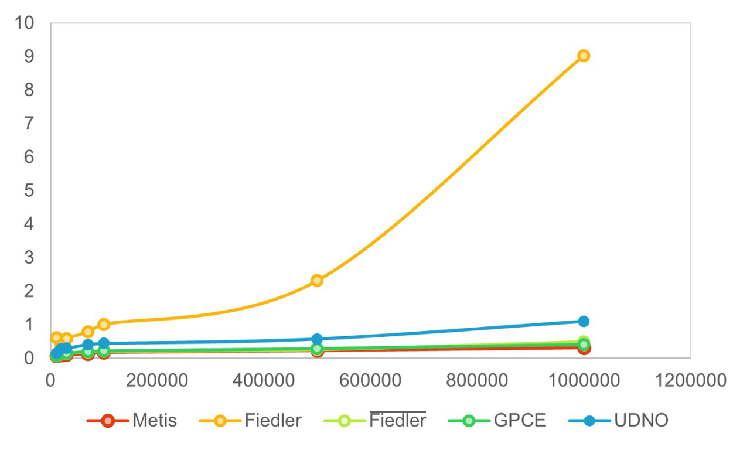}
    }
    \caption{Nnz Ratio and Ordering time changes along with the increase of matrix size in the test collection.}
    \label{fig:scale}
\end{figure}

\subsection{Scalability Analysis}
As shown in Table~\ref{tab:effect}, our test matrix collection is generated from a wide variety of scientific applications. But the training set is only from Delaunay meshes, FEM meshes and 2D and 3D Discretized matrices in SuiteSparse. The prediction performance of UDNO on unseen matrices is also better compared with baselines in Table~\ref{tab:effect}. UDNO is able to be generalized to matrices from other problems.

Sparse matrix size in scientific computing and High Performance Computing (HPC) applications is usually very large. To investigate the effectiveness of our proposed method for extremely large matrices, we illustrate the performance curve changing along with the increase of matrix size in Figure~\ref{fig:scale}(a). For matrix size $n$, we averaged the nnz\_ratios of matrices with size no more than $n$. Besides, we also focus on how the ordering time changes in Figure~\ref{fig:scale}(b). Its curve can be plotted in the same way.

The nnz\_ratio curve changes stably with the increase of matrix size. It increases sublinearly with the matrix size as expected. The ordering time increases linearly with the matrix size. The ordering time of Fiedler vector increases the fastest among all the baselines. However, the spectral embedding method $\overline{\text{Fiedler}}$'s ordering time changes nearly linearly with the matrix size similar to Metis and AMD. UDNO inherits the advantage of spectral embedding model, which captures the global view of the matrix with high computation efficiency.

\section{Conclusion}
We propose an Unsupervised Direct Node Ordering (UDNO) method based on a multi-grid-like GNN architecture, which predicts the ordering of all nodes in a graph simultaneously. First, we employ a multi-level graph neural network architecture that leverages coarsening and interpolation techniques to approximate the eigenvectors corresponding to the first \(d-1\) smallest non-zero eigenvalues of the Laplacian matrix. These resulting eigenvectors are then employed as node features. We introduce a novel loss function based on the 1-sum metric, which quantifies the sum of differences in ordering indices between connected nodes. And then use random variables to make the loss continuous to avoid the inherently discrete and non-differentiable nature of the 1-sum function, allowing us to directly apply gradient descent. This approach effectively circumvents the need for an optimal solution as a target in supervised learning. Experimental results show that our proposed method outperforms traditional graph-theoretic methods and existing deep learning techniques. In the future, we will further conduct in-depth research on deep learning-based sparse matrix reordering methods.

\subsubsection{\ackname}
This research was supported by the National Natural Science Foundation of China under Grant Nos. 62072447 and 12471348 and the National Key R\&D Program of China under Grant No. 2021YFB0300203.

\end{document}